\documentclass[]{interact}
\usepackage{placeins}
\usepackage{epstopdf}
\usepackage{subfigure}
\usepackage[caption=false]{subfig}
\usepackage{natbib}
\bibpunct[, ]{(}{)}{;}{a}{}{,}
\renewcommand\bibfont{\fontsize{10}{12}\selectfont}

\theoremstyle{plain}

\theoremstyle{definition}

\theoremstyle{remark}

\begin{document}

\articletype{Technical Note}

\title{Automatically eliminating seam lines with Poisson editing in complex relative radiometric normalization mosaicking scenarios}

\author{
\name{Shiqi Liu\textsuperscript{a}, Jie Lian\textsuperscript{a}, Xuchen Zhan\textsuperscript{a}, Cong Liu\textsuperscript{a}, Yuze Tian\textsuperscript{a}, Hongwei Duan\textsuperscript{a,*}\thanks{*Corresponding author. Email: duanhw@diit.cn}}
\affil{\textsuperscript{a}Innovation Center of Beijing Data Intelligence Information Technology Co., Ltd., Wuhan, China}
}

\maketitle

\begin{abstract}
Relative radiometric normalization (RRN) mosaicking among multiple remote sensing images is crucial for the downstream tasks, including map-making, image recognition, semantic segmentation, and change detection. However, there are often seam lines on the mosaic boundary and radiometric contrast left, especially in complex scenarios, making the appearance of mosaic images unsightly and reducing the accuracy of the latter classification/recognition algorithms.
This paper renders a novel automatical approach to eliminate seam lines in complex RRN mosaicking scenarios. It utilizes the histogram matching on the overlap area to alleviate radiometric contrast, Poisson editing to remove the seam lines, and merging procedure to determine the normalization transfer order. Our method can handle the mosaicking seam lines with arbitrary shapes and images with extreme topological relationships (with a small intersection area). These conditions make the main feathering or blending methods, e.g., linear weighted blending and Laplacian pyramid blending, unavailable. In the experiment, our approach visually surpasses the automatic methods without Poisson editing and the manual blurring and feathering method using GIMP software.
\end{abstract}


\section{Introduction}

A monthly/quarterly/annually updated sizeable remote sensing image containing the entire region of interest is crucial for downstream change detection tasks, geographic information system, and many other remote sensing applications, e.g., disaster monitoring, environmental monitoring, resource monitoring and geographical mapping. However, limited by the imaging width, mechanism and cloud coverage, a single remote sensing image often cannot cover the entire region of interest. However, suppose several remote sensing images obtained from different satellites, sensors or imaging times are directly mosaicked. In that case, a significant flower-cloth effect will appear (i.e., the mosaic image is not visually continuous.). This phenomenon is since different remote sensing images may be affected by imaging time, illumination effects, atmospheric attenuation, and other factors (\cite{teillet1986image},\cite{kim1990normalization}), resulting in radiometric distortion on the same object. Therefore, eliminating these distortions is important.

\begin{figure}[ht]
\centering
\subfigure[the detailed direct merge result at red point]{%
\resizebox*{3.3cm}{!}{\includegraphics{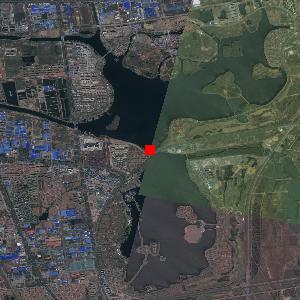}}}
\subfigure[the detailed merge result using histogram matching on overlap area at red point]{%
\resizebox*{3.3cm}{!}{\includegraphics{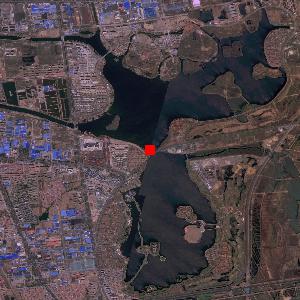}}}
\subfigure[the detailed merge result using histogram matching on overlap area with Poisson editing at red point \textbf{(no seam lines)}]{%
\resizebox*{3.3cm}{!}{\includegraphics{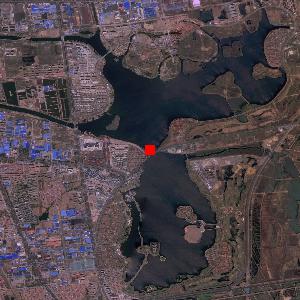}}}
\subfigure[the detailed merge result using histogram matching on overlap area with GIMP manual seam line elimination at red point ]{%
\resizebox*{3.3cm}{!}{\includegraphics{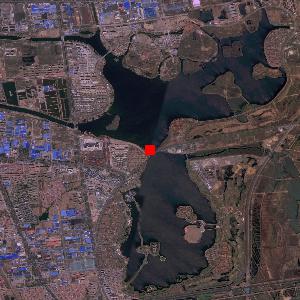}}}
\caption{The comparison figure of the four methods at red point}
\label{fig:Comparison figure of the four methods at red point}
\end{figure}

\begin{figure}[ht]
\centering
\subfigure[the detailed direct merge result at yellow point]{%
\resizebox*{3.3cm}{!}{\includegraphics{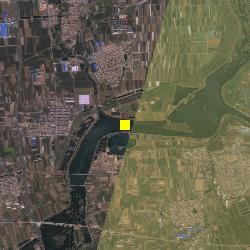}}}
\subfigure[the detailed merge result using histogram matching on overlap area at yellow point]{%
\resizebox*{3.3cm}{!}{\includegraphics{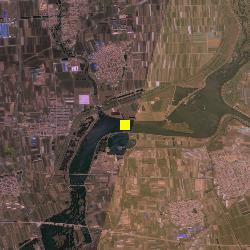}}}
\subfigure[the detailed merge result using histogram matching on overlap area with Poisson editing at yellow point \textbf{(no seam lines)}]{%
\resizebox*{3.3cm}{!}{\includegraphics{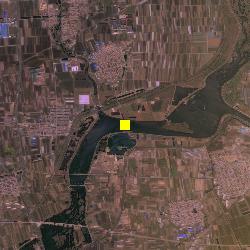}}}
\subfigure[the detailed merge result using histogram matching on overlap area with GIMP manual seam lines elimination at yellow point]{%
\resizebox*{3.3cm}{!}{\includegraphics{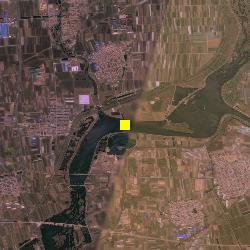}}}
\caption{The comparison figure of the four methods at yellow point}
\label{fig:Comparison figure of the four methods at yellow point}
\end{figure}

Researchers have proposed many methods to alleviate these distortions, but these distortions are not solved effectively because of the seam lines. Relative radiometric normalization(RRN) aims to reduce the global numerical colour/radiometric differences among multiple remote sensing images by modifying the original image's colour/radiometric properties to match the reference image. Wallis filtering (\cite{zhang1999image}, \cite{Li2019Parallel}) is one kind of methods, forcing different regions to have the same mean and variance. Some methods bases on histogram matching (\cite{chavez1994automatic},\cite{horn1979destriping},\cite{weinreb1989destriping}) and some further bases on matching on the overlap area (\cite{helmer2005cloud}). There are also other approaches including pseudo-invariant feature (\cite{schott1988radiometric}), least square regression (\cite{zhang2008automatic}), establishing reference database (\cite{zhang2017mixed}) and etc. However, the methods mentioned above still need post-processing on seam lines to achieve a smooth transition. Besides, in the complex scenarios when there are lots of remote sensing images with different intersecting modes or extreme topological relationships, it becomes hard to use automatic feathering and blending methods to remove seam line. Therefore,  people cost a lot of staffing and time to use Photoshop and other image editing software to manually edit remote sensing images.

Currently, seam lines eliminating methods, including mosaic line optimization, feathering and blending, utilizes overlapped area or vertical/horizontal seam lines shape, failing when the overlapped area is small. For instance, \cite{santellani2018seamless} utilizes Voronoi decomposition and colour contrast cost minimum distance mosaic line optimization to minimize the colour contrast on both sides of the mosaic line. Instead of eliminating the specified seam lines, this kind of methods generates a better mosaic line. However, this kind of methods usually cannot be perfectly seamless and may become ineffective when the overlapped area is small or the overlapped area has inner ring. \cite{lu2014fusion} assumes the rectangular shape of the overlapped area and utilizes the 2-D linear weighted blending on the low frequency overlapped parts, and merges the images based on the mosaic line optimized by the high frequency overlapped parts. However, when the overlapped shape is not rectangular, or the overlapped area is small, this method becomes intractable or unhelpful. Laplacian pyramid (\cite{burt1987Laplacian}) blending utilizes the mask on the overlapped area to construct different levels of blending Gaussian and Laplacian pyramids to reconstruct the blending image to eliminate seam lines. This kind of methods may hopefully overcome arbitrary seam lines on the rectangular overlapped area. However, when the overlapped area is small or complicated, the directly smoothing and downsampling procedures of Laplacian pyramid blending become intractable. Moreover, Laplacian pyramid blending has to change the colour/radiometric of all the images on both sides of seam lines, and this can be a kind of distortion.

This paper presents a novel automatic approach for RRN mosaicking that uses the histogram matching on the overlap area to achieve relative radiometric normalization, Poisson editing to remove the seam line and sequence merging to determine the normalization transfer order. Compared with the traditional seam lines elimination methods, advantages of our method are as follows: \begin{enumerate}
\item \textbf{fully automatic};
\item \textbf{widely applicable} (it can overcome arbitrary seam lines shapes, complex overlapped shapes and small overlapped area.);
\item \textbf{truly seamless with low distortion} (instead of generating a new mosaic line, it achieves seamless transition by eliminating the colour contrast on one side of seam lines, reducing the distortion.).
\end{enumerate}

\section{Methodology}
Suppose that we have a reference image and a set of other images. We can regard the reference image as the starting image of image mosaicking. The proposed method iteratively selects an original image intersected with the reference image and sequentially implements histogram matching, Poisson editing, and merging procedures to generate a new reference image. Repeating the procedures, we will finally obtain the complete mosaicking reference image.

In the following sections, we will first introduce the histogram matching method based on overlap area, and then introduce the Poisson editing on the complex boundary, and finally introduce the merging procedure.

\subsection{Histogram matching based on overlap area}
First, suppose the range of a remote sensing image is $[j_{min},j_{max}]$, we can map it to $[i_{min},i_{max}]$ where $i_{min},i_{max}$ are integers. The density histogram of an image is \begin{equation}\label{density histogram}
  P(i)=\frac{N_i }{N },
\end{equation}
where $N_i$ is the number of pixels that their grey scales equal $i$ and $N$ is the total number of pixels. The cumulative probability of the image is
\begin{equation}\label{cumulative probability}
CP(i) = P(t\le i) = \frac{\sum_{t=1}^{i}N_t}{N}.
\end{equation}
Suppose the overlapped area points set is $S^{overlapped}$, and ${N_t^{overlapped}}_O$ denotes the number of pixels in overlapped original image that grey scales equal $t$ and ${N_t^{overlapped}}_R$ denotes the number of pixels in overlapped reference image that grey scales equal $t$, it yields the cumulative probability of the overlap of the original image and the reference image are $CP_O^{overlapped}$ and $CP_R^{overlapped}$ respectively:
\begin{eqnarray}\label{cumulative probability overlapped}
  CP_O^{overlapped}(i) &=& \frac{\sum_{t=1}^{i}{N_t^{overlapped}}_O}{\vert S^{overlapped}\vert }, \\
  CP_R^{overlapped}(i) &=& \frac{\sum_{t=1}^{i}{N_t^{overlapped}}_R}{\vert S^{overlapped}\vert }.
\end{eqnarray}

For pixel value $i_O$ with cumulative probability $ CP_O^{overlapped}(i_O)$, we can find the corresponding $i_R$, which share the similar probability that
\begin{equation}\label{find cumulative}
i_R = \arg\min_{i}\vert CP_R^{overlapped}(i)- CP_O^{overlapped}(i_O)\vert.
\end{equation}
It yields the mapping $T^{overlap}:i_O\rightarrow i_R$.
We apply the $T^{overlap}$ to the original image then we achieve the histogram matching.

Since objects in the overlap area of the images are supposed consistent, when using the mapping $T^{overlap}$ obtained from the histogram matching on the overlap area of the original image and the reference image, the radiometric value of the processed origin image incline to be consistent to the reference image's radiometric value. Notice that if the reference image and original image only intersect on a small area and the overlapped area's colour/radiometric diversity is insufficient, we skip this procedure and move to the Poisson editing procedure.

However, although the radiometric value can be consistent, there might be noticeable seam lines, especially when the intersection of the original image and reference image is complex or a slight difference occurs in the overlap area. Some feathering and blending methods are only suitable for simple intersection modes such as large rectangle intersection. However, they lose efficacy when the intersection is complex, or the intersection area is small, or the seam lines shapes are complicated, or the radiometric value contrast is too large. An efficient seam lines elimination is needed.

\subsection{Poisson editing on complex boundary}
Poisson editing introduces a guidance vector field from the source image to transplant the source image's gradient vector field into a target region to achieve seamless filling (\cite{perez2003Poisson}).
The central guiding ideology of Poisson editing assumes that seamless image function to be solved in $\Omega\in R^2$ space is $f$, the source image's gradient vector field is $\boldsymbol v$ and the boundary of the seamless image defined on $\partial \Omega$ is $f^*$ where $\partial \Omega$ is the boundary points of $\Omega$. The task is to find the optimal $f^{optimal}$ such that
\begin{equation}\label{Poisson function}
  f^{optimal}=\arg\min\limits_{f}\int\int_{\Omega}\Vert \nabla f -\boldsymbol v\Vert^2 \quad \mathrm{with} \ f\vert_{ \partial\Omega} = f^*\vert_{ \partial \Omega},
\end{equation}
where $\Vert \cdot \Vert$ is the $L_2$ norm and $\nabla$ is the gradient operator. The minimizing problem is equivalent to solve:
\begin{equation}\label{Poisson function}
  f^{optimal}=\arg\min\limits_{f}\triangle f =\mathrm{div} \ \boldsymbol v \ \mathrm{over} \ \Omega \quad \mathrm{with} \ f\vert_{ \partial\Omega} = f^*\vert_{ \partial \Omega},
\end{equation}
where $\triangle$ is the laplace operator and $\mathrm{div}$ is divergence operator.

The above equations are generalized to the discrete case (linear equations) to make the Poisson editing algorithm trackable, and we can further solve it by iteration methods. The detailed can refer to \cite{perez2003Poisson}.

The important thing left is to determine the source point set $\Omega$ and the boundary function $f^*$ on $\partial \Omega$. The following shows how to obtain the $\Omega$ and $f^*$ on $\partial \Omega$.

First, we have the processed original image $I_O$ with point sets $S_O=\{\boldsymbol p|\boldsymbol p \in origin\ image\}$ and the reference image $I_R$ with point set $S_R=\{\boldsymbol p|\boldsymbol p \in reference\ image\}$. The example is illustrated in the following figure~\ref{fig:The illustration of the reference and origin point set}.
\begin{figure}[ht]
\centering
\subfigure[reference point set $S_R$]{%
\resizebox*{4cm}{!}{\includegraphics{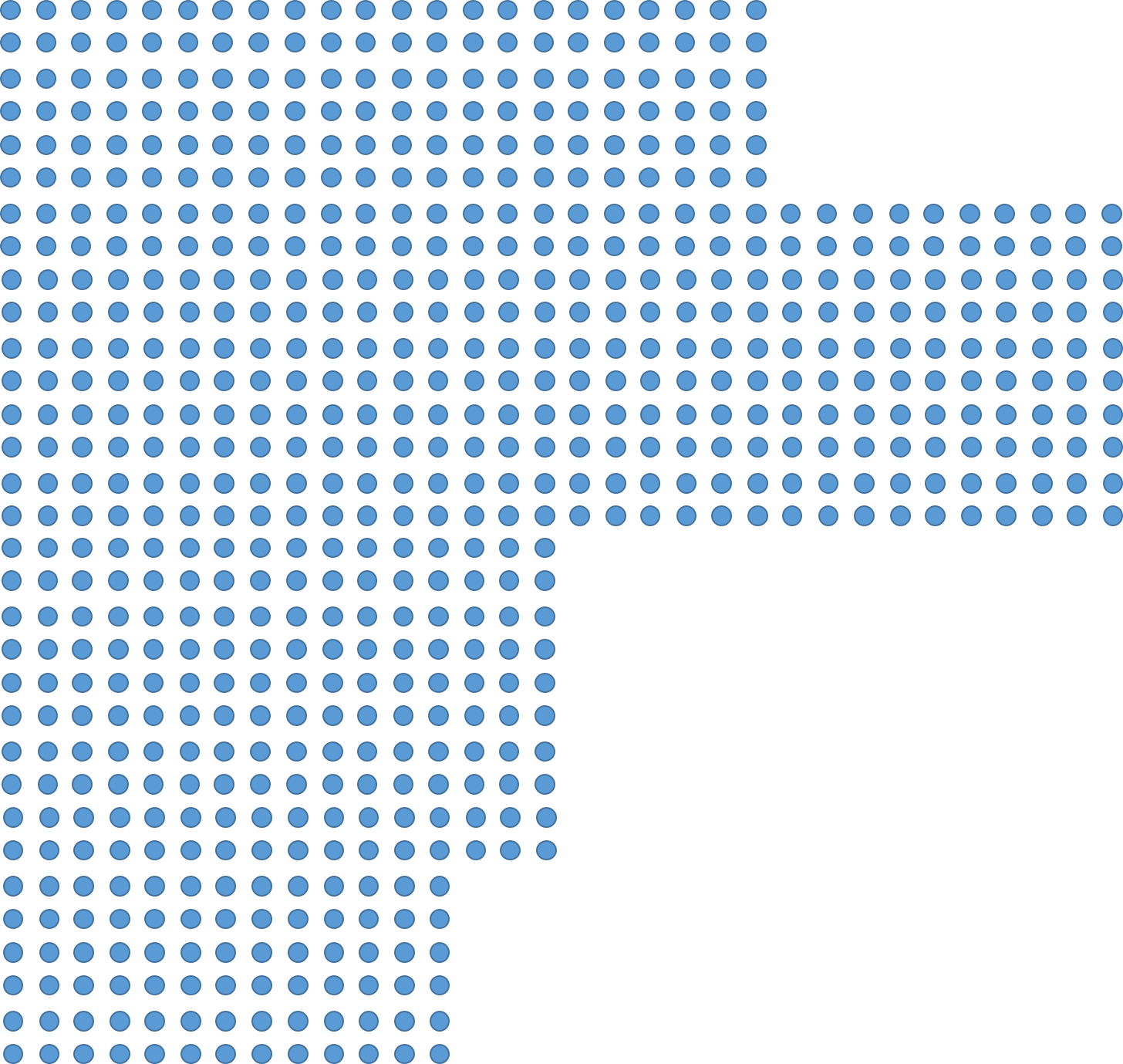}}}\hspace{5pt}
\subfigure[original point set $S_O$]{%
\resizebox*{4cm}{!}{\includegraphics{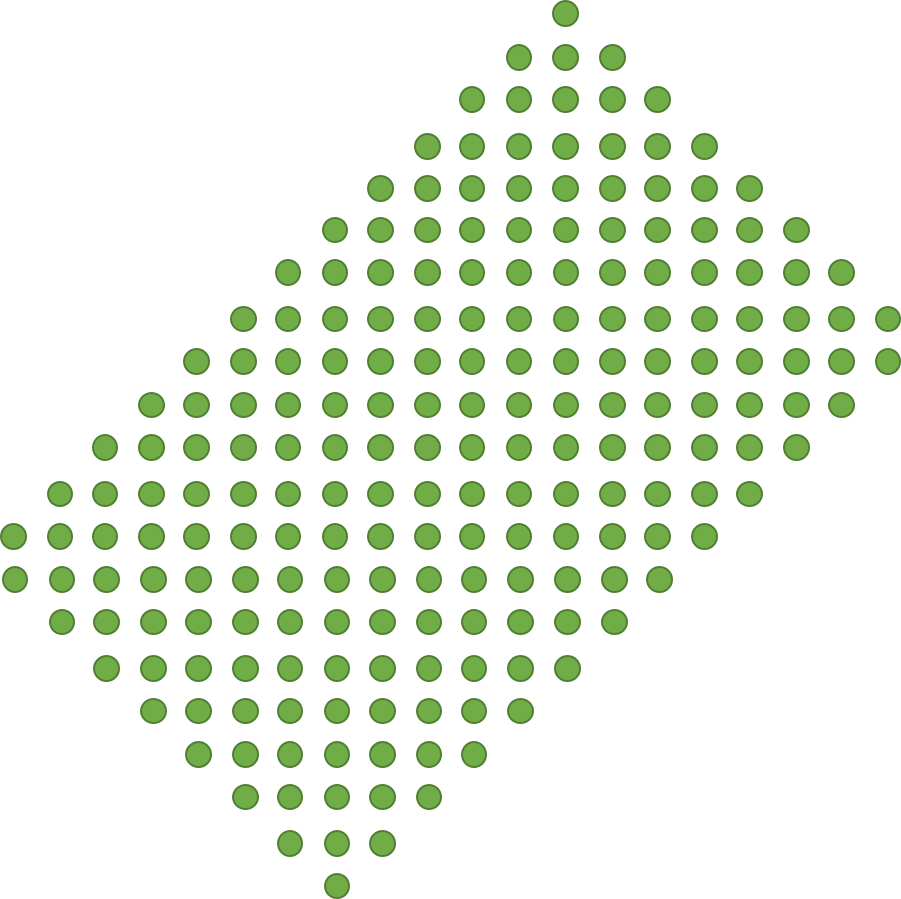}}}\hspace{5pt}
\subfigure[whole point set $S_O\cup S_R$]{%
\resizebox*{4cm}{!}{\includegraphics{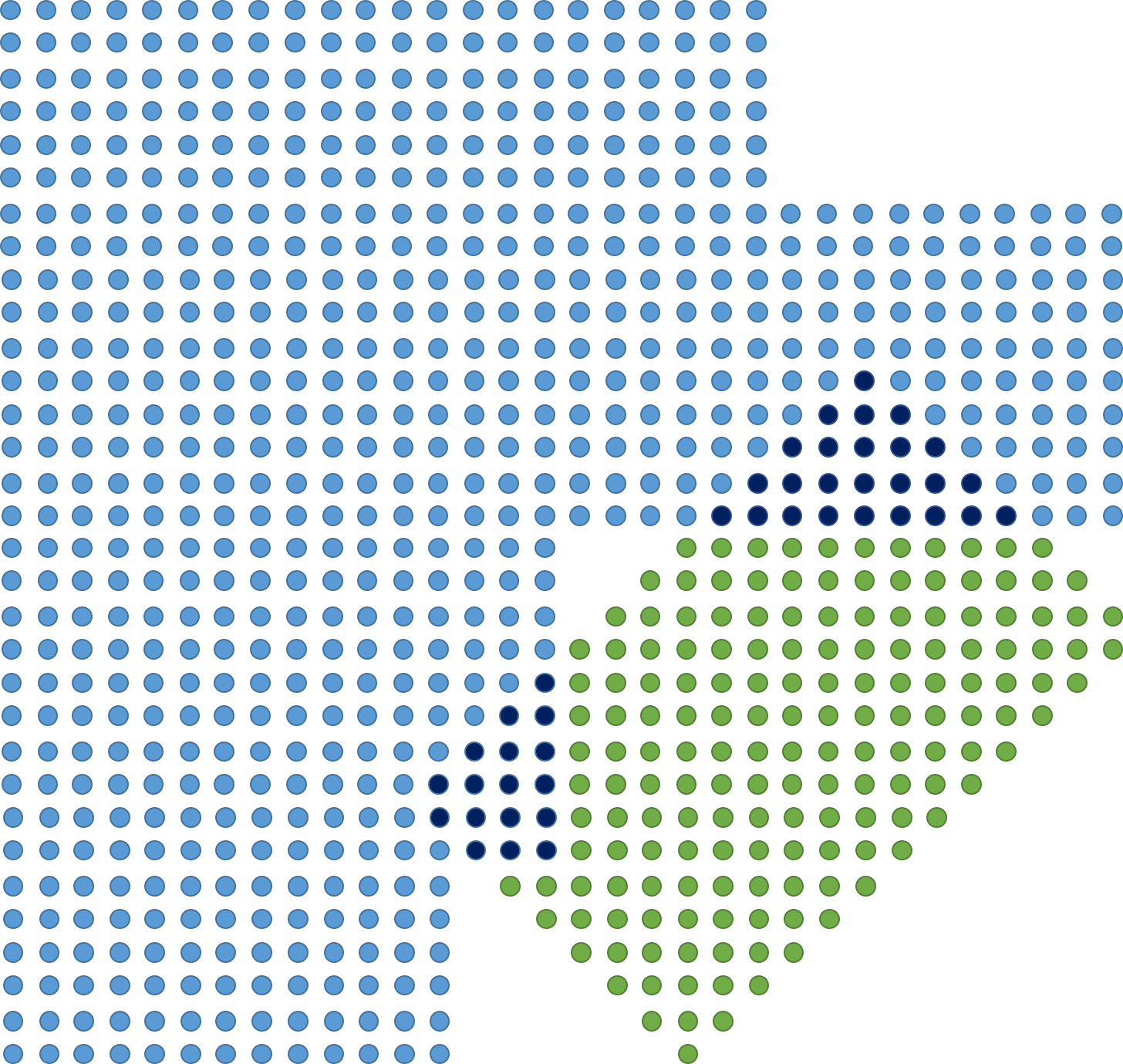}}}
\caption{The illustration of the reference and original point set. The dark blue points are the intersection points.} \label{fig:The illustration of the reference and origin point set}
\end{figure}

Second, the target image $I_T$ is with the region $S_T = S_O-S_R^\circ$ where $-$ is the set difference and $S_R^\circ$ is the interior point of $S_R$. By the interior point of $S_R$, we mean that \begin{equation}\label{interior point}
S_R^\circ=\{\boldsymbol q|\boldsymbol q\in S_R ,\ Neighbor(\boldsymbol q)\in S_R \ \mathrm{and} \ \vert Neighbor(\boldsymbol q)\vert =4\}
     \end{equation}
where $Neighbor(\boldsymbol q)$ is the set of $\boldsymbol q$'s 4-connected neighbors. Besides,
\begin{equation}\label{target image}
  I_T(p)=\left\{
            \begin{aligned}
I_O(p) \ \boldsymbol p \in  (S_O-S_R^\circ)-\partial S_R\\
I_R(p) \ \boldsymbol p \in   (S_O-S_R^\circ)\cap \partial S_R
\end{aligned}
\right.
\end{equation}
where $\partial S_R = S_R - S_R^\circ$.
\begin{figure}[ht]
\centering
\subfigure[the reference interior point set $S_R^\circ$]{%
\resizebox*{4cm}{!}{\includegraphics{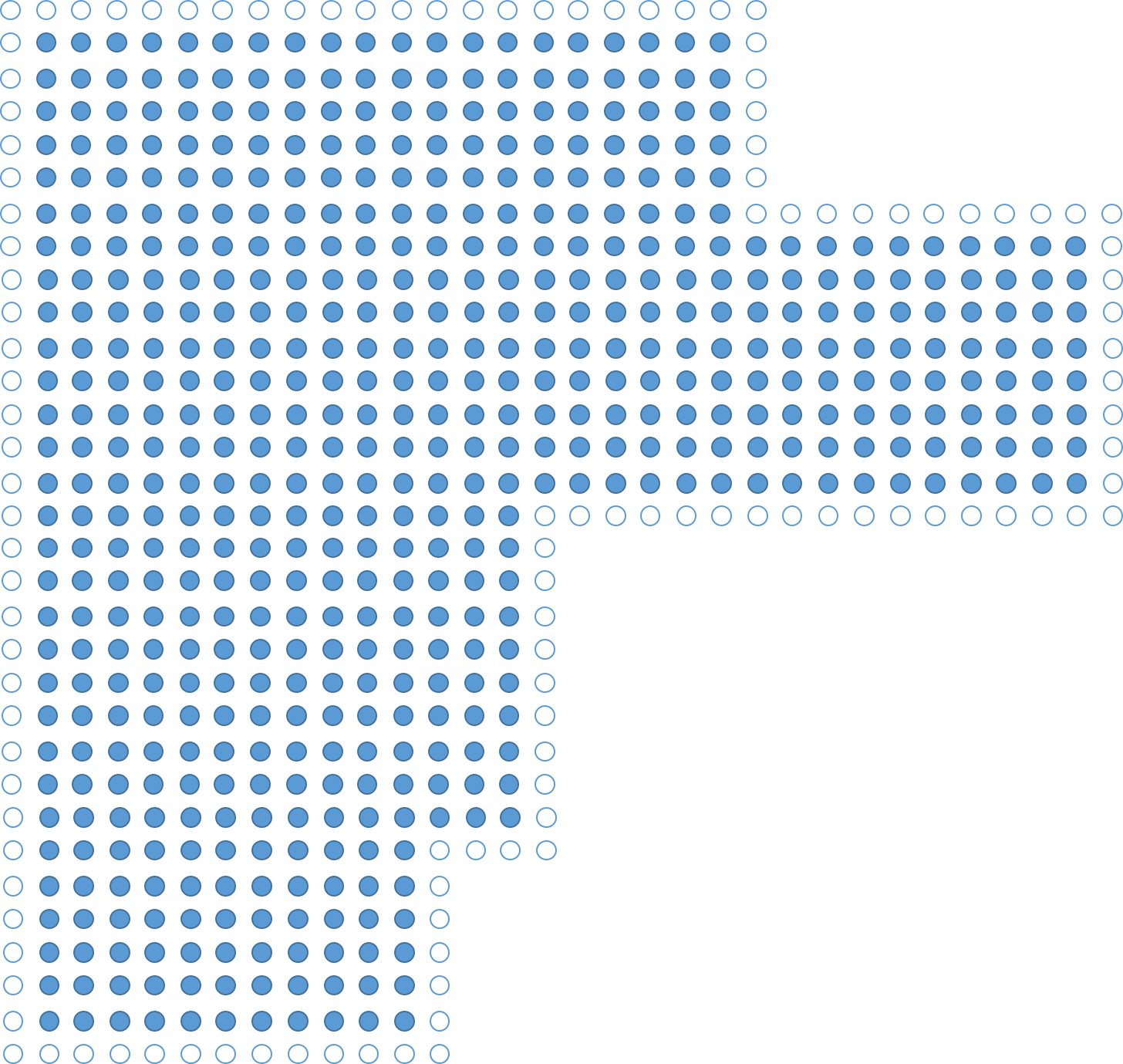}}}\hspace{5pt}
\subfigure[the target point set $S_T= S_O-S_R^\circ$]{%
\resizebox*{4cm}{!}{\includegraphics{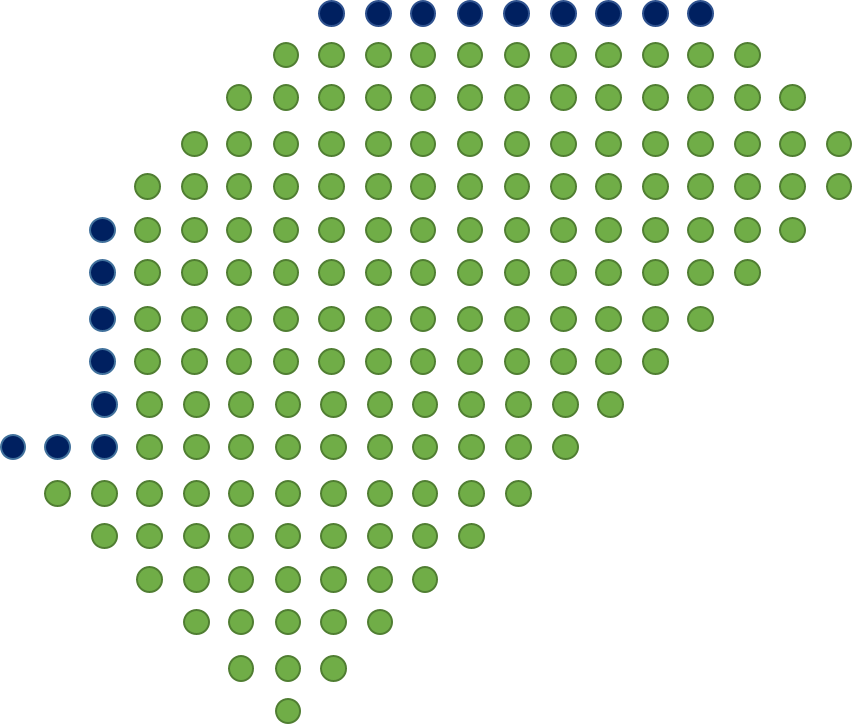}}}\hspace{5pt}
\subfigure[the source point set $S_S= S_T^\circ$]{%
\resizebox*{4cm}{!}{\includegraphics{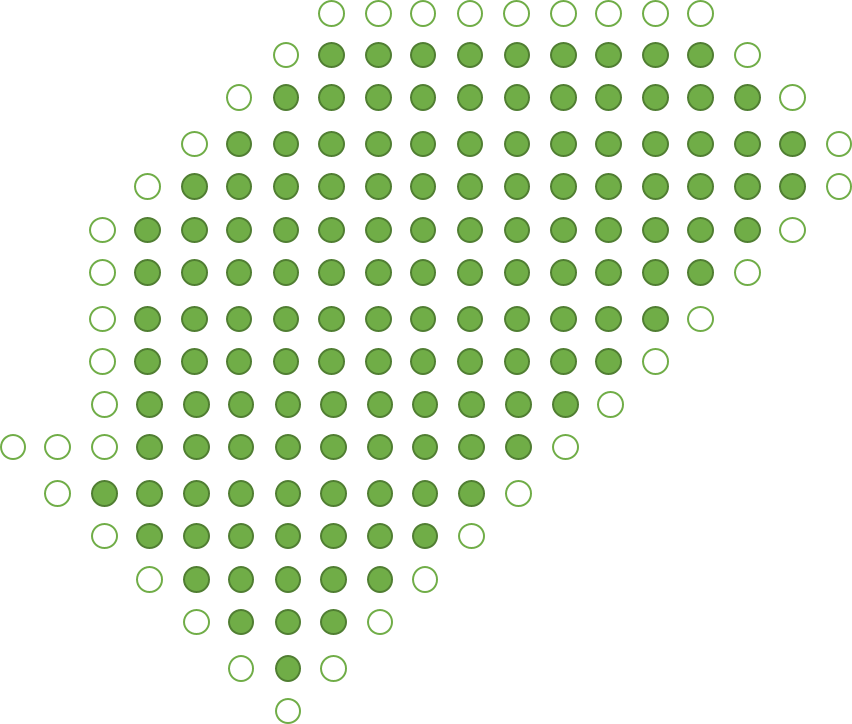}}}
\caption{The illustration of the reference interior, target and source point set. The dark blue points are the intersection points. The white points which don't belong to the point set demonstrate original set's boundary.}
\label{fig:The illustration of the reference interior, target and source point set}
\end{figure}

The source image $I_S$ is with the region $S_S=S_T^\circ$ and $I_S={I_O}_{\vert S_S}$, that is $I_S(\boldsymbol p)=I_O(\boldsymbol p)\ \forall \boldsymbol p\in S_S$.
It is obvious that the source image is contained in the target image, and the target image is modified slightly on the boundary point of the $S_O-S_R^\circ$. The point value of $I_R$ on $(S_O-S_R^\circ)\cap \partial S_R$ replaces of $I_O$ on $(S_O-S_R^\circ)\cap \partial S_R$. Notice that replacement introduce the seam line between the boundary and the interior points. The example of the aforementioned point set is illustrated in the figure~\ref{fig:The illustration of the reference interior, target and source point set}.

Finally, let $\Omega = S_T$ and $f^*\vert_{\partial \Omega} = I_T\vert_{\partial S_T}$, we can implement the Poisson editing to transplant source image to the target image and obtained the seamless image where the color contrast of the boundary and interior points are eliminated due to the guided interpolation.

\begin{figure}[ht]
\centering
\subfigure[the new source point set $S_S^{new}$]{%
\resizebox*{4cm}{!}{\includegraphics{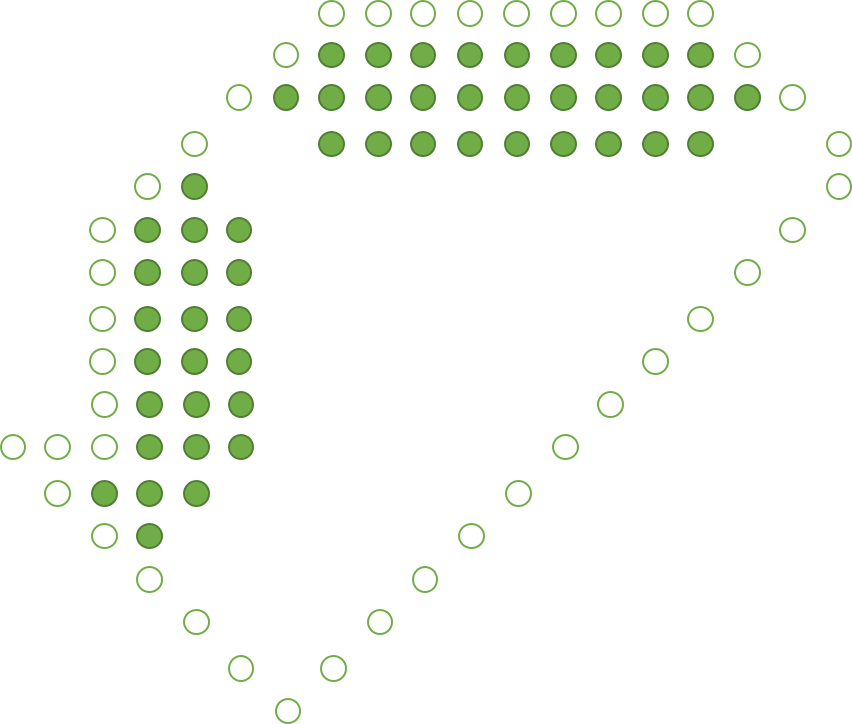}}}
\subfigure[the new target point set $S_T^{new}$]{%
\resizebox*{4cm}{!}{\includegraphics{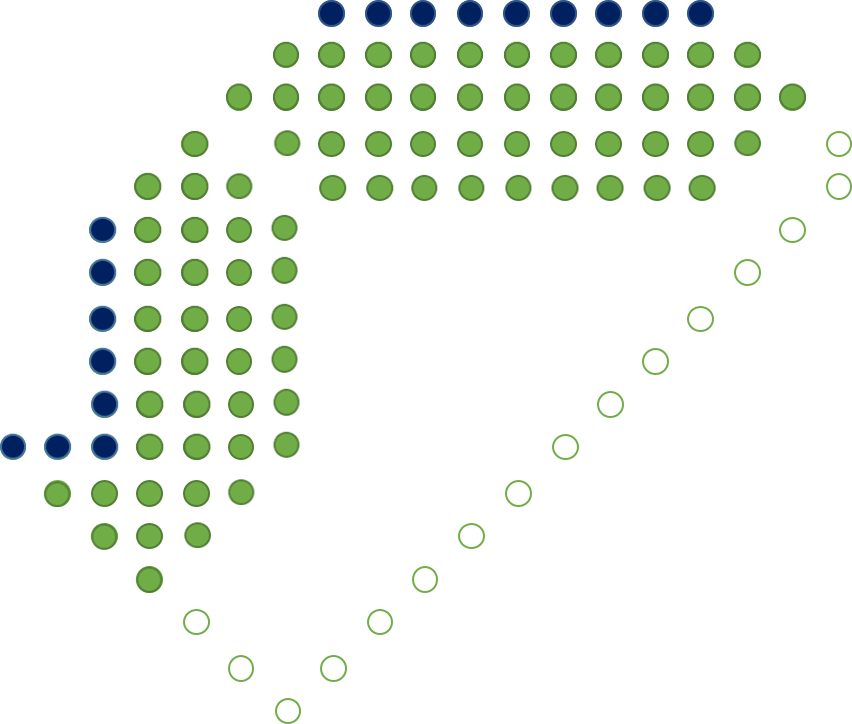}}}
\caption{The illustration of the new source point set with $d=3$. The white points which don't belong to the point set demonstrate original set's boundary. The dark blue points are the intersection points.}
\label{fig:new source point set}
\end{figure}

Notice that the size of source image $I_S$ might be pretty large, and it will cost many resources to solve the Poisson equation. Empirically,  only the pixel value of the region in $S_S$ that near  $(S_O-S_R^\circ)\cap \partial S_R$ will be changed drastically after the Poisson editing. Therefore, we consider reducing the size of $I_S$ in the actual implementation. Let \begin{equation}\label{Sphere of a point}
  B_d(\boldsymbol p)=\{\boldsymbol q|\vert q_1 -p_1\vert +\vert q_2-p_2\vert \le d\},
\end{equation}
and
\begin{equation}\label{Sphere of a set}
   B_d(S)=\cup_{\boldsymbol p \in S}   B_d(\boldsymbol p) .
\end{equation}
We build a new source image $I_S^{new}$ with $S_S^{new} = S_S \cap B_d(S_T\cap \partial S_R)$  and $S_T^{new} = S_T \cap B_{d+1}(S_T\cap \partial S_R)$. Let $\Omega = S_T^{new}$ and $f^*\vert_{\partial \Omega} = I_T\vert_{\partial S_T^{new}}$ implement the Poisson editing of the new source image to the target image. The choice of $d$ depends on the size of $I_O$ and depends on whether we implement the histogram matching procedure. If we skip the histogram matching procedure, $d$ should be set to the maximum value to also eliminate global color contrast. The $S_S^{new}$ is illustrated in the figure~\ref{fig:new source point set}.

\subsection{Merging}
\begin{figure}[ht]
  \centering
  \includegraphics[width=4cm]{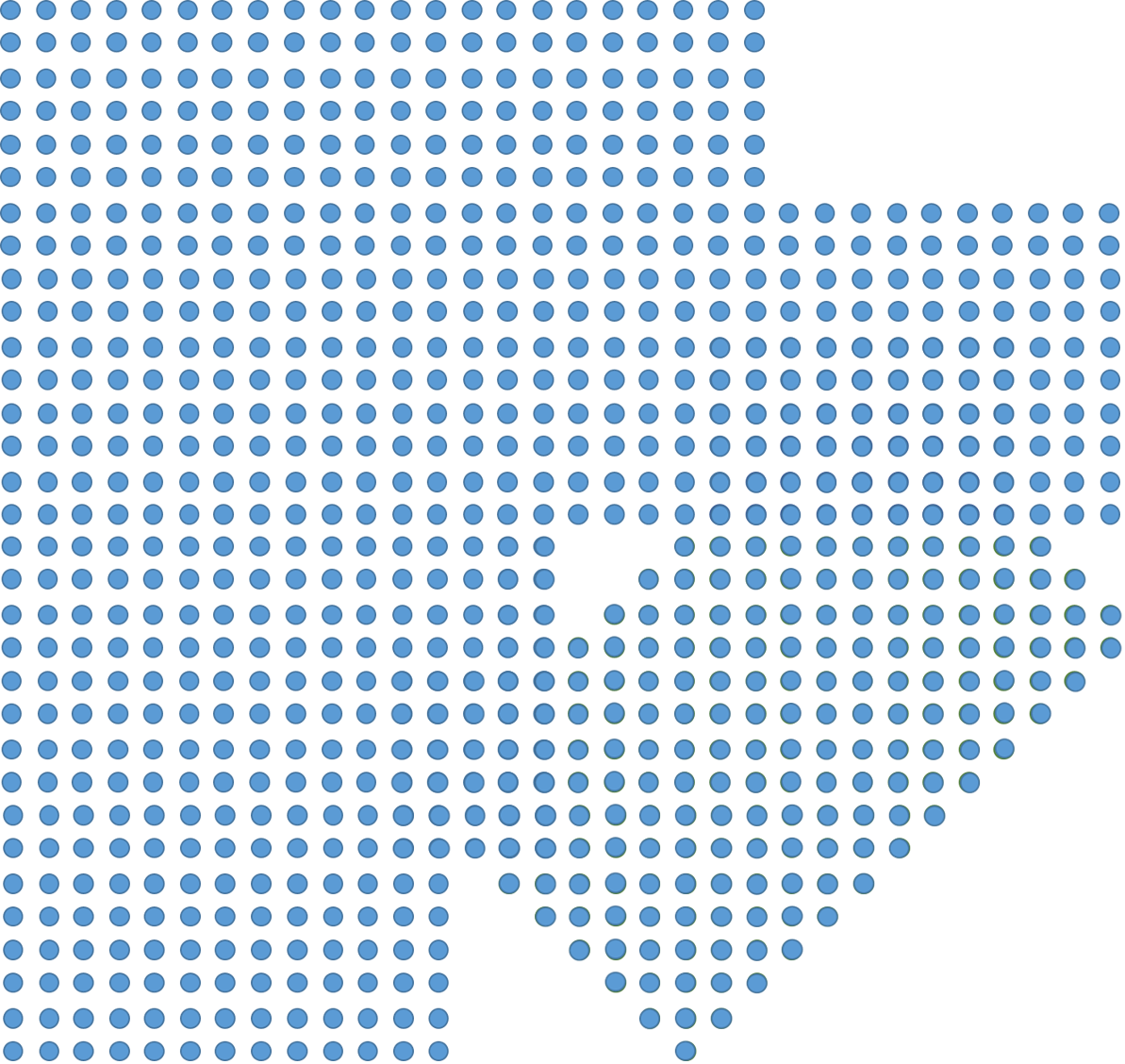}
  \caption{the merged new reference image}\label{new reference image}
\end{figure}

In order to let this procedure continue, we merge the seamless target image and the reference image. Notice that since the seam lines are removed, the merged image tends to be seamless and can be the new reference image shown in the figure~\ref{new reference image} to participate in the latter repeated processing.

\FloatBarrier
\section{Experiments}
\begin{figure}[ht]
\centering
\subfigure[the direct merge result]{%
\resizebox*{7cm}{!}{\includegraphics{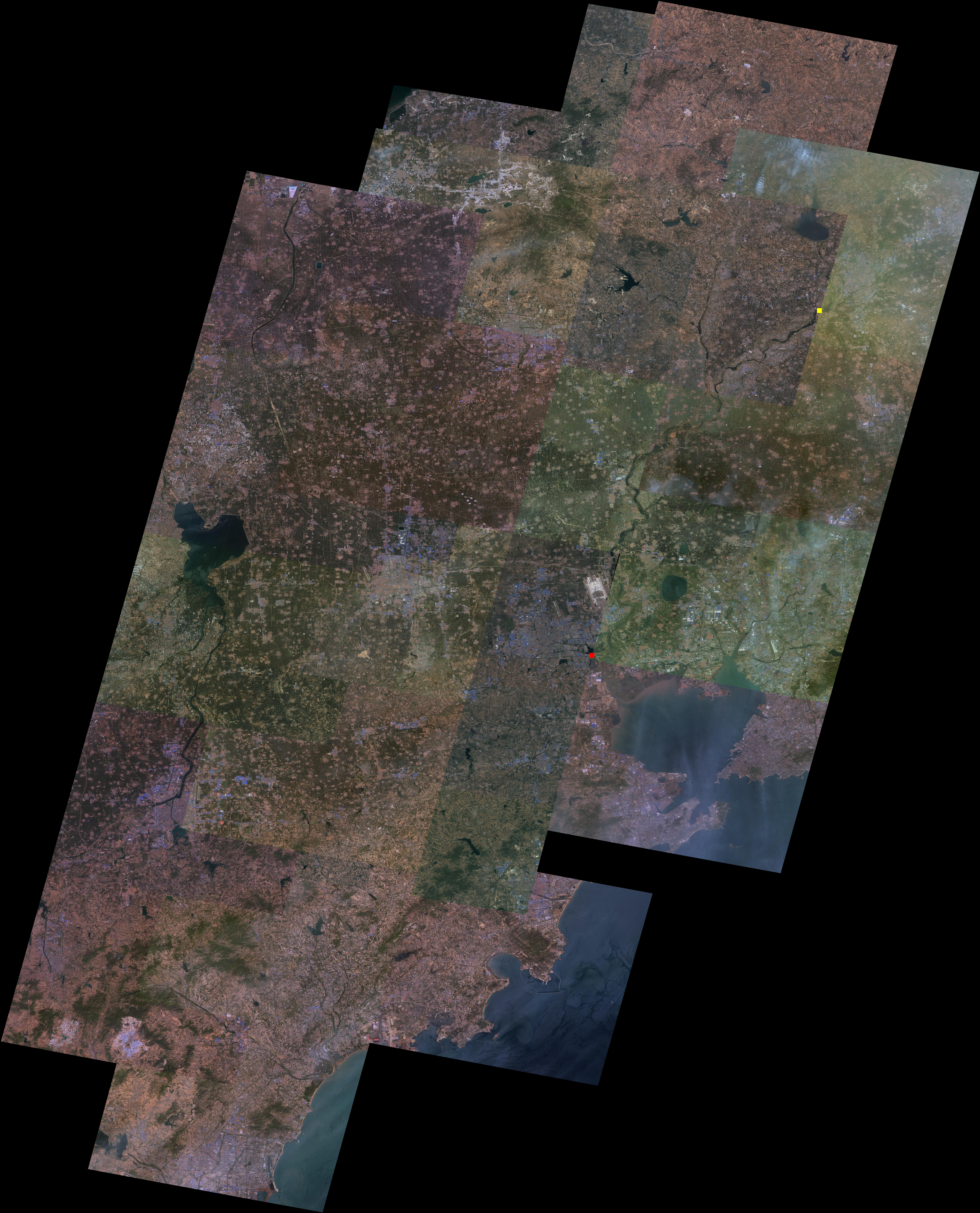}}}
\subfigure[the merge result using histogram matching on overlap area]{%
\resizebox*{7cm}{!}{\includegraphics{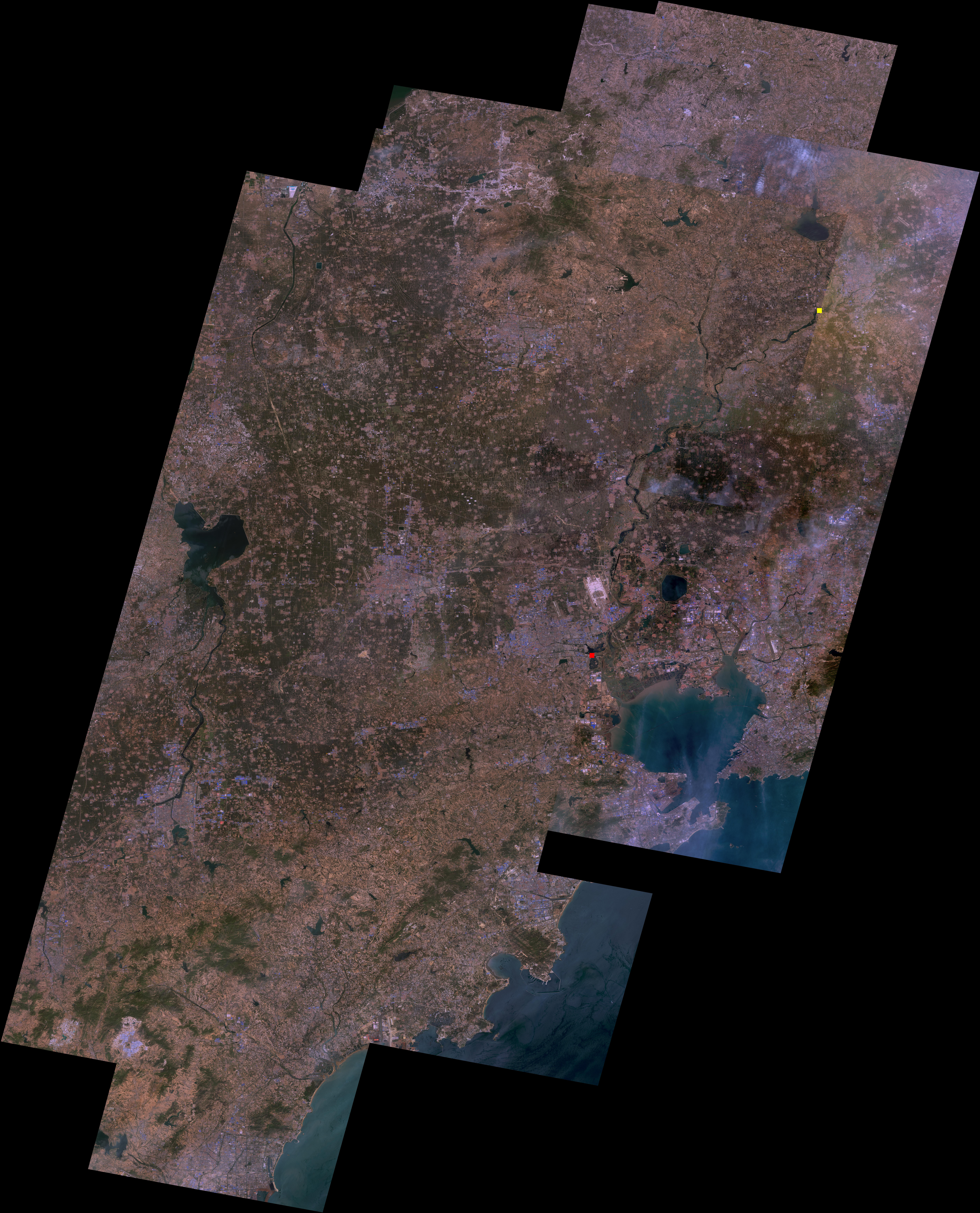}}}\hspace{5pt}
\subfigure[the merge result using histogram matching on overlap area with Poisson editing]{%
\resizebox*{7cm}{!}{\includegraphics{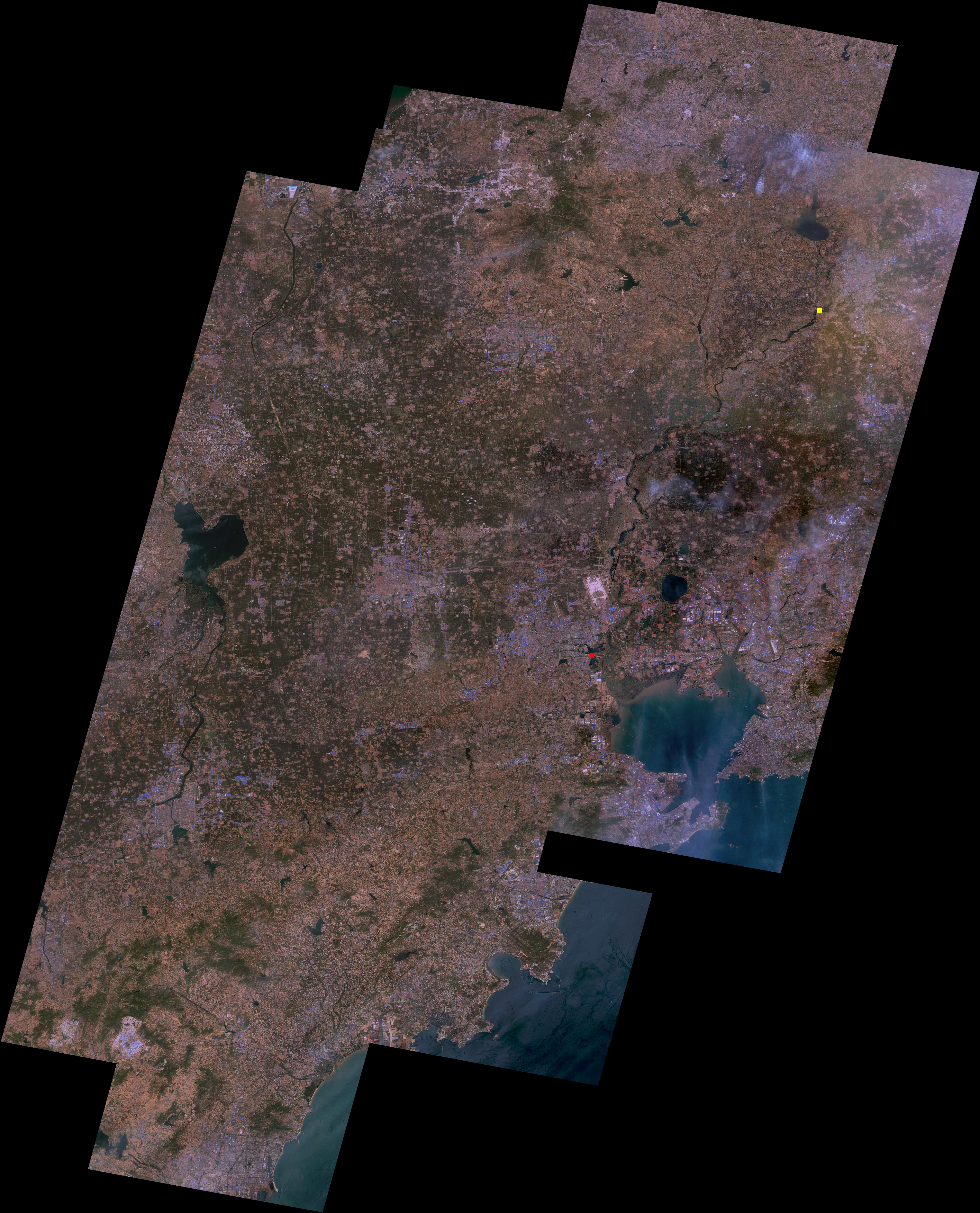}}}
\subfigure[the merge result using histogram matching on overlap area with GIMP manual seam line elimination]{%
\resizebox*{7cm}{!}{\includegraphics{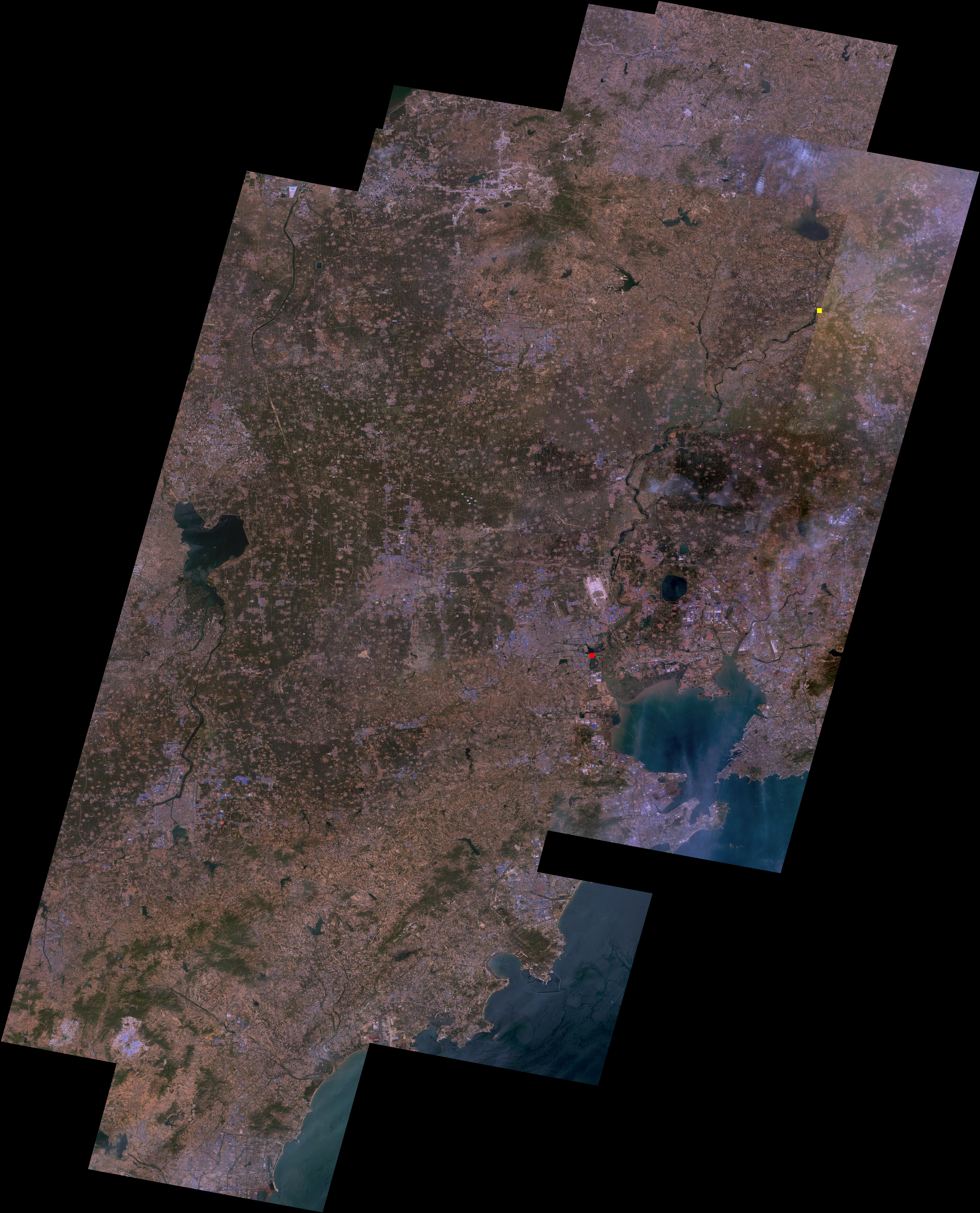}}}
\caption{The comparison figure of the four methods  }
\label{fig:Comparison figure of the four methods}
\end{figure}
\subsection{Material and environment}
We implement the algorithm and apply 34 resampled remote sensing multispectral images of GF-1 and GF-2 satellites in image mosaicking task.

These images partially cover the range of Qingdao, Shandong Province of China. These images were acquired in May 2019.

We performed the preprocessing steps of RPC orthorectication, radiometric calibration, atmospheric correction and  Gram-Schmidt PanSharpening (\cite{laben2000process}) on these images, and finally resampled the spatial resolution to  16m. The first three spectral are displayed.

We implement the algorithm on the computer with Inter(R) Xeon(R) W-2123 3.60GHz CPU  and 32G RAM.

\subsection{Image mosaicking}

We use three automatic methods and one manual method with the help of GIMP image editing software(\cite{gimp}). We use python to accomplish the automatic algorithm. Precisely, we first calculate the intersection of the square envelope of the reference image and the square envelope of the original image to speed up the calculation of the overlapped area and the determination of the blue boundary points($S_T\cap\partial S_R$). The parameter $d$ defined in equation~\ref{Sphere of a point} is set to  $150$ in this experiment. Then we use the sparse matrix of scipy library(\cite{2020SciPy}) to construct the Poisson matrix and coefficient and use the algebraic multi-grid method with pyamg(\cite{OlSc2018}) library to solve the Poisson equations. We use gdal library to merge the images.

The time-consuming algebraic multi-grid solving step of  Poisson editing on about $4000-750000$ points costs about $1-21$ second. The average processing time on single remote sensing image ranging from about $(2000\times1500)$ to $(2900\times2400)$ is about 1.5 minute.

Figure~\ref{fig:Comparison figure of the four methods} demonstrates the direct mosaic result, histogram matching result, histogram matching with Poisson editing result and histogram matching with GIMP manual seam line elimination. From figure~\ref{fig:Comparison figure of the four methods}, we can tell from the figure that histogram matching eliminates the colour contrast.

We provide the detailed image of the red point and yellow point in the figure~\ref{fig:Comparison figure of the four methods at red point} and figure~\ref{fig:Comparison figure of the four methods at yellow point}.

According to figure~\ref{fig:Comparison figure of the four methods at red point} and \ref{fig:Comparison figure of the four methods at yellow point}, histogram matching on overlap area cannot deal with the boundary seam lines in the complex situation, including the boundary lake and multiple intersection areas. The colour difference on the different parts of the intersection area can cause colour inconsistency after histogram matching. The first case happened in the area shown in the figure~\ref{fig:Comparison figure of the four methods at yellow point}, and these two cases happened together in the area shown in the figure~\ref{fig:Comparison figure of the four methods at red point}. The upper-right remote sensing image in the figure~\ref{fig:Comparison figure of the four methods at red point} has more intersection with the image above rather than the intersection of image on the left, while the colour is different on these two images. As a result, it causes the colour difference on the lake. The manual edition blurs and makes the feathering effect on the image and alleviate the seam lines. In contrast, the Poisson editing can eliminate the seam lines on the complex situation and spread the colour to the expansion region and make the transition smooth. Compare to the four methods,  the histogram matching on overlap area with Poisson editing method has the best effect.
\FloatBarrier
\section{Conclusion}

This paper presents a novel automatic approach for eliminating seam lines in RRN mosaicking tasks. It utilizes the histogram matching on the overlap area to achieve colour/radiometric normalization, Poisson editing to remove the complicated seam lines in diverging intersecting modes and merging to achieve seamless mosaicking finally. It is fully automatic, overcoming arbitrary seam lines shapes, complex intersecting modes and small overlapped area. Our method visually surpasses the automatic methods without Poisson editing in the experiment and the manual blurring and feathering method using GIMP software.

Currently, our method is independent of the mosaic line optimization algorithm, and it can easily incorporate with the optimized mosaic line. Besides, our method also naturally applies to cloud inpainting task. In the future, we will investigate how to accelerate the Poisson editing procedure to let this algorithm be more efficient than the current version.

\section*{Acknowledgements}

We would like to thank Shengbing Zou's guidance and leadership. We would like to thank Jian Liao for discussion the Poisson editing algorithm. We would like to thank Fan Wang for discussing the computation of the overlapped area.

\section*{Disclosure statement}

None.

\bibliographystyle{tfcad}

\begin{thebibliography}{19}
\newcommand{\enquote}[1]{``#1''}
\providecommand{\natexlab}[1]{#1}
\providecommand{\url}[1]{\normalfont{#1}}
\providecommand{\urlprefix}{}

\bibitem[Burt and Adelson(1987)]{burt1987Laplacian}
Burt, Peter~J, and Edward~H Adelson. 1987. ``The Laplacian pyramid as a compact
  image code.'' In \emph{Readings in computer vision}, 671--679. Elsevier.

\bibitem[Chavez and MacKinnon(1994)]{chavez1994automatic}
Chavez, Pat~S, and David~J MacKinnon. 1994. ``Automatic detection of vegetation
  changes in the southwestern United States using remotely sensed images.''
  \emph{Photogrammetric engineering and remote sensing} 60 (5).

\bibitem[Helmer and Ruefenacht(2005)]{helmer2005cloud}
Helmer, Eileen~H, and Bonnie Ruefenacht. 2005. ``Cloud-free satellite image
  mosaics with regression trees and histogram matching.'' \emph{Photogrammetric
  Engineering \& Remote Sensing} 71 (9): 1079--1089.

\bibitem[Horn and Woodham(1979)]{horn1979destriping}
Horn, Berthold~KP, and Robert~J Woodham. 1979. ``Destriping Landsat MSS images
  by histogram modification.'' \emph{Computer Graphics and Image Processing} 10
  (1): 69--83.

\bibitem[Kim and Elman(1990)]{kim1990normalization}
Kim, Hongsuk~H, and Gregory~C Elman. 1990. ``Normalization of satellite
  imagery.'' \emph{International Journal of Remote Sensing} 11 (8): 1331--1347.

\bibitem[Laben and Brower(2000)]{laben2000process}
Laben, Craig~A, and Bernard~V Brower. 2000. ``Process for enhancing the spatial
  resolution of multispectral imagery using pan-sharpening.'' Jan.~4. US Patent
  6,011,875.

\bibitem[Lu, Li, and Fu(2014)]{lu2014fusion}
Lu, Ting, Shutao Li, and Wei Fu. 2014. ``Fusion based seamless mosaic for
  remote sensing images.'' \emph{Sensing and Imaging} 15 (1): 1--14.

\bibitem[Olson and Schroder(2018)]{OlSc2018}
Olson, L.~N., and J.~B. Schroder. 2018. ``{PyAMG}: Algebraic Multigrid Solvers
  in {Python} v4.0.'' Release 4.0,
  \urlprefix\url{https://github.com/pyamg/pyamg}.

\bibitem[P{\'e}rez, Gangnet, and Blake(2003)]{perez2003Poisson}
P{\'e}rez, Patrick, Michel Gangnet, and Andrew Blake. 2003. ``Poisson image
  editing.'' In \emph{ACM SIGGRAPH 2003 Papers}, 313--318.

\bibitem[Santellani, Maset, and Fusiello(2018)]{santellani2018seamless}
Santellani, Emanuele, Eleonora Maset, and Andrea Fusiello. 2018. ``SEAMLESS
  IMAGE MOSAICKING VIA SYNCHRONIZATION.'' \emph{ISPRS Annals of Photogrammetry,
  Remote Sensing \& Spatial Information Sciences} 4 (2).

\bibitem[Schott, Salvaggio, and Volchok(1988)]{schott1988radiometric}
Schott, John~R, Carl Salvaggio, and William~J Volchok. 1988. ``Radiometric
  scene normalization using pseudoinvariant features.'' \emph{Remote sensing of
  Environment} 26 (1): 1--16.

\bibitem[Shuo et~al.(2019)]{Li2019Parallel}
Shuo, Li, Hui Wang, Liyong Wang, Xiangzhou Yu, and Le~Yang. 2019. ``Parallel
  color balancing method using adaptive block Wallis algorithm for image
  mosaicking.'' \emph{Journal of Remote Sensing} 23 (04): 706--716.

\bibitem[Teillet(1986)]{teillet1986image}
Teillet, PM. 1986. ``Image correction for radiometric effects in remote
  sensing.'' \emph{International Journal of Remote Sensing} 7 (12): 1637--1651.

\bibitem[{The GIMP Development Team}(2019)]{gimp}
{The GIMP Development Team}. 2019. ``GIMP.''
  \urlprefix\url{https://www.gimp.org}.

\bibitem[Virtanen et~al.(2020)]{2020SciPy}
Virtanen, Pauli, Ralf Gommers, Travis~E. Oliphant, Matt Haberland, Tyler Reddy,
  David Cournapeau, Evgeni Burovski, et~al. 2020. ``SciPy 1.0: Fundamental
  Algorithms for Scientific Computing in Python.'' \emph{Nature Methods} .

\bibitem[Weinreb et~al.(1989)]{weinreb1989destriping}
Weinreb, MP, R~Xie, JH~Lienesch, and DS~Crosby. 1989. ``Destriping GOES images
  by matching empirical distribution functions.'' \emph{Remote Sensing of
  Environment} 29 (2): 185--195.

\bibitem[Zhang et~al.(2008)]{zhang2008automatic}
Zhang, L, L~Yang, H~Lin, and Mingsheng Liao. 2008. ``Automatic relative
  radiometric normalization using iteratively weighted least square
  regression.'' \emph{International Journal of Remote Sensing} 29 (2):
  459--470.

\bibitem[Zhang, Zhang, and Zhang(1999)]{zhang1999image}
Zhang, Li, Zuxun Zhang, and Jianqing Zhang. 1999. ``The image matching based on
  wallis filtering.'' \emph{Journal of Wuhan Technical University of Surveying
  and Mapping} 24 (1): 24--27.

\bibitem[Zhang et~al.(2017)]{zhang2017mixed}
Zhang, Yongjun, Lei Yu, Mingwei Sun, and Xinyu Zhu. 2017. ``A mixed radiometric
  normalization method for mosaicking of high-resolution satellite imagery.''
  \emph{IEEE Transactions on Geoscience and Remote Sensing} 55 (5): 2972--2984.

\end{thebibliography}

\end{document}